\def\year{2019}\relax
\documentclass[letterpaper]{article} 
\usepackage{aaai19}  
\usepackage{times}  
\usepackage{helvet}  
\usepackage{courier}  
\usepackage{url}  
\usepackage{graphicx}  
\usepackage{amsmath}
\usepackage{tabularx}
\usepackage{multirow}
\usepackage{bm}
\begin{document}
%
\title{DeRPN: Taking a further step toward more general object detection}
\author{Lele Xie, Yuliang Liu, Lianwen Jin$^*$, Zecheng Xie\\
School of Electronic and Information Engineering, South China University of Technology\\
xie.lele@mail.scut.edu.cn, liu.yuliang@mail.scut.edu.cn, $^*$eelwjin@scut.edu.cn, zchengxie@gmail.com\\
}
\nocopyright
\maketitle
\begin{abstract}
Most current detection methods have adopted anchor boxes as regression references. However, the detection performance is sensitive to the setting of the anchor boxes. A proper setting of anchor boxes may vary significantly across different datasets, which severely limits the universality of the detectors. To improve the adaptivity of the detectors, in this paper, we present a novel dimension-decomposition region proposal network (DeRPN) that can perfectly displace the traditional Region Proposal Network (RPN). DeRPN utilizes an anchor string mechanism to independently match object widths and heights, which is conducive to treating variant object shapes. In addition, a novel scale-sensitive loss is designed to address the imbalanced loss computations of different scaled objects, which can avoid the small objects being overwhelmed by larger ones. Comprehensive experiments conducted on both general object detection datasets (Pascal VOC 2007, 2012 and MS COCO) and scene text detection datasets (ICDAR 2013 and COCO-Text) all prove that our DeRPN can significantly outperform RPN. It is worth mentioning that the proposed DeRPN can be employed directly on different models, tasks, and datasets without any modifications of hyperparameters or specialized optimization, which further demonstrates its adaptivity. The code will be released at https://github.com/HCIILAB/DeRPN.
\end{abstract}

\section{Introduction}
\noindent Recently, general object detection has achieved rapid development, driven by the convolutional neural network (CNN). As a significant task in computer vision, general object detection is expected to detect more object classes \cite{redmon2016yolo9000,singh2017r} and perform impressively on different datasets \cite{everingham2010pascal,lin2014microsoft}. Unfortunately, we notice that most of the general object detection methods are not very general. When employing some state-of-the-art methods \cite{Dai2017Deformable,lin2018focal,zhang2018single} on different datasets, it is usually indispensable to redesign the hyperparameters of regression references, termed as anchor boxes in Region Proposal Network (RPN) \cite{ren2015faster}. Even for some specific object detection tasks, such as scene text detection \cite{lyu2018multi,zhang2017feature,TextBoxes2018}, directly applying state-of-the-art approaches of general object detection cannot produce their due effects \cite{liao2017textboxes,liu2017deep}. This problem stems from presupposed anchor boxes with fixed shapes and scales.
\begin{figure}[t]
\centering
\includegraphics[width=3.2in]{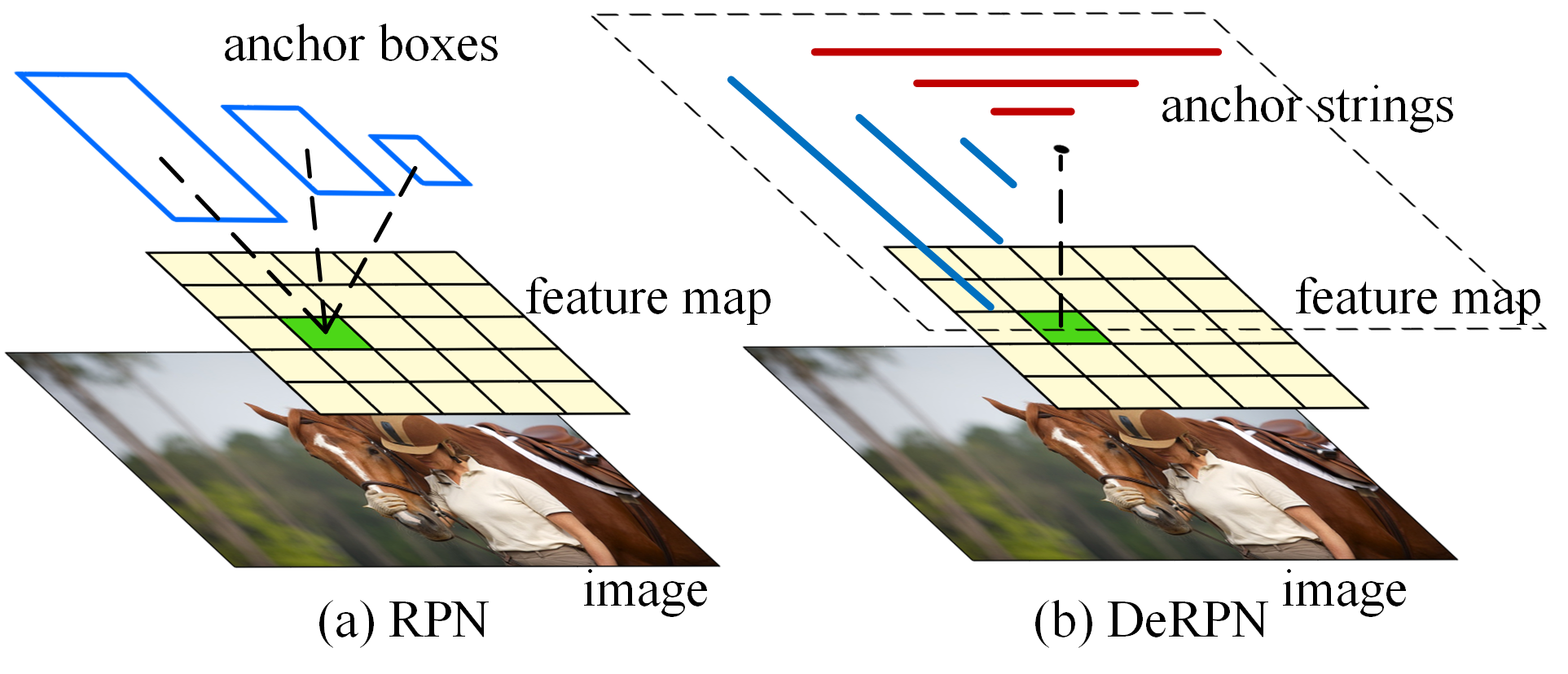}
\caption{Schemas of RPN and DeRPN: (a) RPN adopts multiple anchor boxes of fixed shapes and scales, (b) DeRPN divides bounding boxes into flexible anchor strings, decoupling width and height. }
\label{predicted_vector}
\end{figure}
Although RPN was verified to be an effective method to generate region proposals in \cite{ren2015faster}, the anchor boxes adopted in RPN are very sensitive, which limits the adaptivity to variant objects. Once the anchor boxes excessively deviate from the ground truths in a dataset, the performance is curtailed significantly. This is why many specific methods for scene text detection \cite{liao2017textboxes,Ma2017Arbitrary,zhang2017feature} lay emphasis on the design of the anchor boxes, which are intended to be long, narrow, and consistent with the text characteristics. However, manually setting the anchor boxes is cumbersome and it is difficult to guarantee the best performance. Although \cite{redmon2016yolo9000} proposed an algorithm based on K-means clustering to precompute anchor boxes, the improvement was still limited.

As illustrated in Fig.1 (a), RPN adopts anchor boxes as regression references. We argue that anchor boxes of fixed scales and shapes have extreme difficulty covering thousands of objects. Therefore, we propose a novel dimension-decomposition region proposal network (DeRPN). In Fig.1 (b), DeRPN divides anchor boxes into several independent segments that we call anchor strings, inheriting the terminology from \cite{ren2015faster}. Note that DeRPN decomposes the detection dimension by decoupling the width and height. The main contributions of our work are summarized as follows:
\begin{itemize}
\item We propose a novel region proposal network, DeRPN, which has strong adaptivity. Without any modifications for hyperparameters, DeRPN can be directly employed on different models, tasks, or datasets.
\item The proposed DeRPN adopts an advanced dimension-decomposition mechanism. Through flexible anchor strings, DeRPN can match objects with optimal regression references, which allows the network to be more smoothly trained.
\item We propose a novel scale-sensitive loss function to tackle the imbalance of object scales and prevent small objects from being overwhelmed by lager ones.
\item The proposed DeRPN maintains a consistent network and running time with RPN, and thus it can conveniently be transplanted to current two-stage detectors.
\item The proposed DeRPN has a higher recall rate and more accurate region proposals than the previous RPN. It outperforms RPN on different tasks and datasets.
\end{itemize}

\section{Related Works}
In the past several years, early object detection methods were confined to traditional methods \cite{viola2001rapid,felzenszwalb2010object} involving handcrafted features. With the recent significant progress of CNN, more CNN-based methods have been put forward and rapidly dominated the detection task. We introduce some of the works below.

\subsubsection{General object detection}Object detection methods are categorized into two streams: two-stage approaches and one-stage approaches. The two-stage framework contains two procedures, where the first step is to generate some candidate regions of interest (RoIs) with region proposal methods, e.g., Selective Search (SS) \cite{uijlings2013selective} and RPN \cite{ren2015faster}. Then, to determine the accurate regions and clear classes further, the second step resamples features according to the RoIs with a pooling operation \cite{girshick2015fast,dai2016r}. Such two-stage methods \cite{lin2016feature,he2017mask,li2018r} have achieved high detection accuracy on some challenging datasets. Conversely, the one-stage methods \cite{redmon2016yolo9000,lin2018focal,zhang2018single} focus more on the running speed. Without the resampling operation, the procedure of one-stage methods has been simplified manyfold and thus benefits the running speed.

\subsubsection{Region proposal methods}As an important component in two-stage detectors, region proposal methods have a significant impact on the final detection result. Formerly, correlative image processing methods were used to generate region proposals. Some were based on grouping pixels \cite{uijlings2013selective}, and others were based on sliding windows \cite{zitnick2014edge}. Note that these methods are independent of the detectors. Later, \citeauthor{ren2015faster} proposed a region proposal network that was integrated in Faster R-CNN to build an end-to-end detector. Notably, as proven by \cite{ren2015faster}, RPN  outperformed other previous proposal methods, for example SS \cite{uijlings2013selective} and EdgeBoxes \cite{zitnick2014edge}. To date, the vast majority of two-stage detectors have adopted RPN to generate region proposals.

\subsubsection{Scene text detection}Different from general objects, scene texts are usually smaller, thinner, and with characteristic texture and rich diversity in their aspect ratios \cite{tian2015text}. In addition, scene text detection can be applied to many scenarios, for example multilingual translation, document digitization, and automatic driving. For these reasons, scene text detection has been regarded as another challenging task in computer vision. Comprehensive survey of scene text detection can be found in \cite{ye2015text}. According to the basic detection element, methods of scene text detection are classified into three categories: (1) character based \cite{hu2017wordsup}, (2) word based \cite{deng2018pixellink}, and (3) text line based \cite{zhang2016multi}.

\section{Methodology}
In this section, we describe the details of the DeRPN. The network and full pipeline are illustrated in Fig. 2.
\begin{figure*}[t]
\label{pipeline}
\centering
\includegraphics[width=5.in]{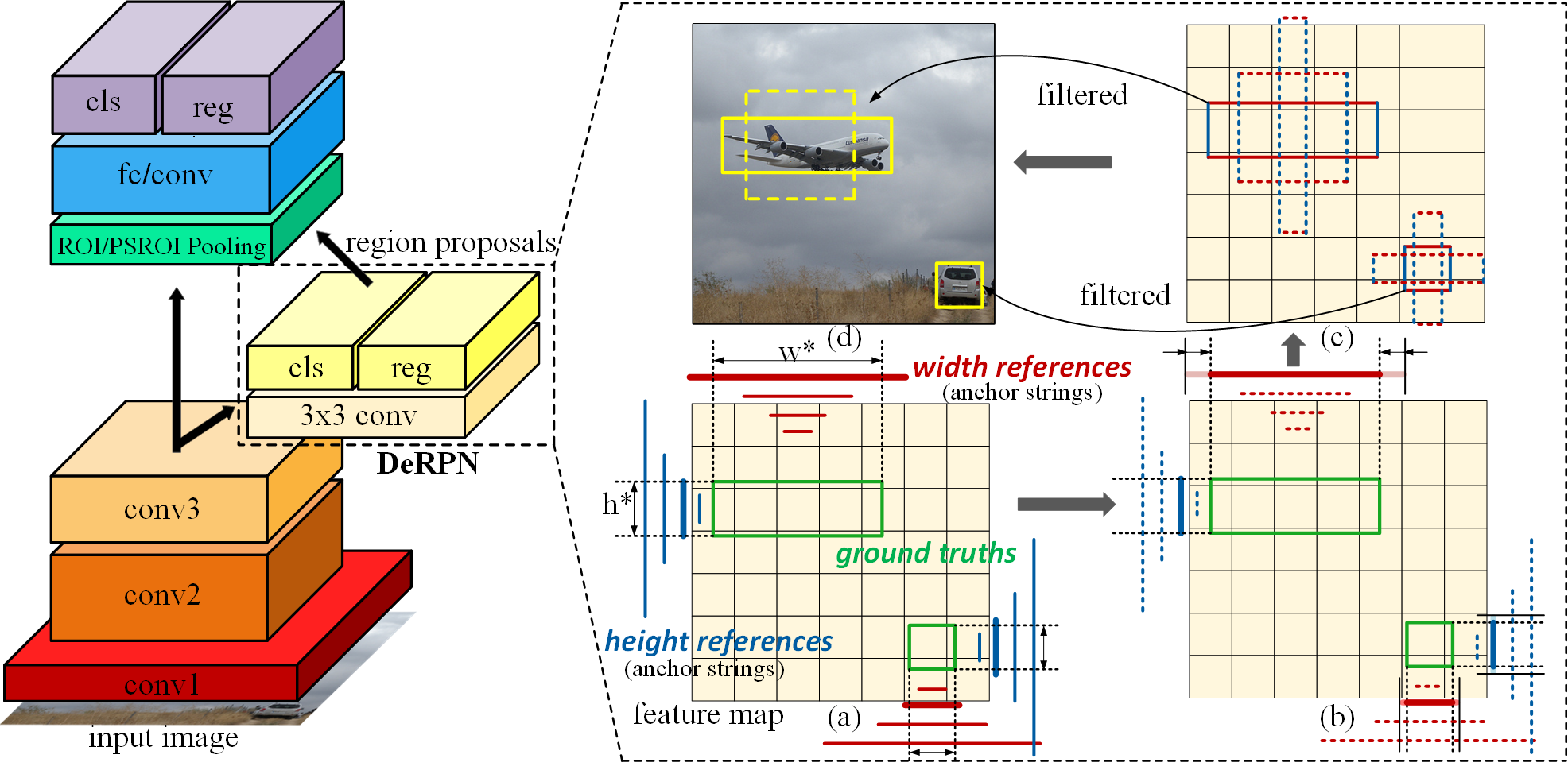}
\caption{DeRPN network and pipeline. (a) Object widths and heights are independently matched with anchor strings. Bold lines represent well-matched anchor strings. This step is employed during the training phase. (b) We apply classification and regression on anchor strings. Dashed lines are anchor strings of low probability. (c) Predicted width and height segments are combined to compose bounding boxes. (d) We filter bounding boxes by probability and NMS to generate region proposals.}
\end{figure*}
\subsection{Modeling}
Most current methods regard the object detection as adjustment (reg) and filtration (cls) toward complete bounding boxes. Based on a convolutional neural network, these methods take the features ($\bm{x}$) from a CNN and input them to a regression layer and classification layer. Usually, the regression layer, implemented by a convolutional or fully connected layer, is a linear operation to predict the parameterized coordinates ($\bm{t}$). To acquire the predicted bounding boxes, these parameterized coordinates are decoded according to anchor boxes ($B_a$). In addition, the classification layer applies an activation function (e.g. Sigmoid or Softmax, denoted as $\sigma$) on the predicted values to generate the probability ($P_B$) of bounding boxes. The mathematical descriptions of this procedure are as follows:
\begin{align}
\label{rpn_des}
\bm{t}&=\bm{{\rm{W_r}}}\bm{x}+\bm{b_r}\\
B(x,y,w,h)&=\psi(\bm{t},B_a(x_a,y_a,w_a,h_a))\\
P_B&=\sigma(\bm{{\rm{W_c}}}\bm{x}+\bm{b_c})
\end{align}
where $\bm{{\rm{W_r}}}$ and $\bm{b_r}$ denote the weights and biases of the regression layer. Similarly, $\bm{{\rm{W_c}}}$ and $\bm{b_c}$ are those for the classification layer. $x$, $y$, $w$, and $h$ are coordinates of the bounding boxes. $\psi$ represents an anti-parameterized function that is used to decode coordinates. \\
\indent However, thousands of objects possess extremely variant shapes, and harshly setting numerous anchor boxes proves to be adverse for training and very time-consuming \cite{ren2015faster,redmon2016yolo9000,liu2017deep}. The underlying problem is that people are difficult to design appropriate anchor boxes to treat diverse objects. The excessive deviations between anchor boxes and ground truths will aggravate the learning burden of detectors and dramatically curtail performance. A novel solution is to decompose the detection dimension by decoupling the width and height to alleviate the impact from variant shapes of objects. With the dimension-decomposition mechanism, we introduce anchor strings ($S_a^w(x_a,w_a)$, $S_a^h(y_a,h_a)$) that serve as independent regression references for the object width and height. Differently, anchor strings predict independent segments ($S_w(x,w)$,$S_h(y,h)$) and corresponding probabilities ($P_s^w$,$P_s^h$), instead of full bounding boxes. We describe the procedure as follows:
\begin{align}
\label{drpn_des1}
\bm{t}^w\!\!=&\bm{{{\rm{W_r}}}}^w\bm{x}\!\!+\!\bm{b_r}^w&S_w(x,w)\!=&\psi(\bm{t}^w\!,S_a^w(x_a,\!w_a))\\
\bm{t}^h\!\!=&\bm{{\rm{W_r}}}^h\bm{x}\!\!+\!\bm{b_r}^h&S_h(y,h)\!=&\psi(\bm{t}^h\!,S_a^h(y_a,\!h_a)))\\
P_s^w\!\!=&\sigma(\bm{{\rm{W_c}}}^w\bm{x}\!+\!\bm{b_c}^w)&P_s^h\!=&\sigma(\bm{{\rm{W_c}}}^h\bm{x}\!+\!\bm{b_c}^h)
\end{align}
\indent As the two-dimensional bounding boxes are required for the detection results, we need to reasonably combine the predicted segments to recover the bounding boxes. The combination procedure is given by
\begin{align}
\label{drpn_des2}
B(x,y,w,h)=~&f(S_w(x,w), S_h(y,h)) \\
P_B = ~&g(P_s^w,P_s^h)
\end{align}
where, $f$ represents a kind of rule or algorithm to combine predicted segments. Meanwhile, $g$ is a function (e.g., arithmetic mean, harmonic mean) which evaluates the probabilities of combined bounding boxes.

Through discretization, we assume that the object width or height in a dataset has $n$ kinds of changes. Fully combining all widths and heights determines approximately $n^2$ object shapes with which the anchor box needs to be faced. In other words, the matching complexity of the anchor box is $O(n^2)$. Nevertheless, when adopting the dimension-decomposition mechanism, $n$ kinds of widths and heights are independently matched with anchor strings, which produces a lower matching complexity of $O(n)$.
\subsection{Dimension Decomposition}
\subsubsection{Anchor strings}
The previous RPN regarded anchor boxes as its regression references. For the difficulty of matching various objects with fixed bounding boxes, RPN heavily relies on the design of anchor boxes and further loses satisfactory adaptivity. By contrast, DeRPN breaks the two-dimensional boxes into a series of independent one-dimensional segments called anchor strings. Through this dimension decomposition, we separately handle the object width and height for classification and regression, which can alleviate the impact from diverse object shapes.

Anchor strings are expected to cover widths and heights of all objects. We set the anchor strings as a geometric progression (denoted as $\{a_n\}$), i.e., (16, 32, 64, 128, 256, 512, 1024). Theoretically, this geometric progression can apply to widths or heights within a large range of [$8\sqrt2$, $1024\sqrt2$], which can cover most of the objects in many scenarios. In this paper, both the width and height use the same but independent geometric progression as their anchor strings. As illustrated in Fig. 2 (a), these anchor strings are assigned at each feature map location. Each anchor string is responsible for predicting the width segment or height segment, instead of a full bounding box.

The next concern is how to choose the best-matched anchor strings for an object. In RPN, an anchor box is chosen based on the intersection over union (IoU) between the anchor box and ground truth. If the IoU is over 0.7, or the IoU is the largest one, that anchor box is regarded as positive. Owing to the deviation between the anchor box and object, sometimes the IoU is very small (less than 0.2). Under this situation, the anchor box is probably still chosen for its largest IoU, which produces a significant regression loss in the training state. Conversely, as shown in Fig. 2 (a), DeRPN reasonably matches the objects with anchor strings based on length instead of IoU. The best-matched anchor strings are evaluated by
\begin{equation*}
M_j\!=\!\{i|\arg\min\limits_{i}\left|\log e_j\!-\!\log a_i\right|\}
\cup\{i,i+1|\left|\frac{e_j}{a_i}\!-\!\sqrt q\right|\!\leq\!\beta\}
\end{equation*}
\begin{equation}
\label{eq_match}
~~~~~~~~~~~~~~~~~~~~~~~~~~~~~~~~~~~~~~~~~~~~~~(i=1,2,3,...,N)
\end{equation}
\noindent where $M_j$ denotes the index set of matched anchor strings for the $j$-th object. $e_j$ is the object edge (width or height). $N$ and $q$ represent the number of terms, and the common ratio (in this paper, $q$ is set to 2) of geometric progression $\{a_n\}$, respectively. Term $a_i$ is $i$-th anchor string in $\{a_n\}$. Note that the first term in this equation represents the closest anchor string to edge $e_j$. The second term describes a transition interval within $[(\sqrt q-\beta)\!\times\! a_i, (\sqrt q+\beta)\!\times\! a_i]$, where $\beta$ is used to adjust magnitude of interval, which is intended to be 0.1 in our experiments. If $e_j$ locates in the transition interval, both $i$ and $i+1$ are chosen as matched indexes. We employ the transition interval to reduce ambiguities resulting from factors such as image noises and ground truth deviations.

Notably, the anchor string mechanism of DeRPN introduces boundedness that ensures a stable training procedure. By neglecting the transition interval, it is easy to prove that the largest deviation (measured by ratio) between the anchor string and object edge is at most $\sqrt{q}$, which means the regression loss of DeRPN is bounded. Compared to RPN, the unexpected small IoU usually generates a significant regression loss. Empirically, RPN cannot even converge if the anchor boxes deviate excessively from the ground truths.

\subsubsection{Label assignment}
To assign labels, initially, we define aligned anchor strings that locate at the object centers on a feature map. The positive labels are assigned to aligned anchor strings if they are well matched with corresponding objects, as identified with Eq. (\ref{eq_match}). Except for the aligned ones, we employ an observe-to-distribute strategy to determine other anchor strings. That is, we first observe the regression results for each anchor string. After regression, the predicted segments (widths or heights) of the anchor strings are combined to compose region proposals. If the region proposals have high IoUs (over 0.6) with ground truths, we distribute positive labels to the corresponding anchor strings. The detailed combination procedure is introduced in the next section. Anchor strings that do not satisfy the above conditions will simply be treated as negative.

The previous RPN statically assigns labels merely based on the IoU. Owing to deviations between the anchor boxes and objects, sometimes the IoUs are very small, which introduces a huge regression loss. Compared with RPN, DeRPN has proposed a new kind of secure, dynamic, and reasonable mechanism for label assignment. In most cases, the features at the object centers are representative, and thus such a label assignment method for aligned anchor strings is reasonable. Except for the centers, we cannot identify whether the features at other positions are important. Consequently, we adopt a dynamic observe-to-distribute strategy to determine the labels at other positions conservatively.

\subsubsection{Consistent network}
DeRPN has maintained the consistent network architecture with RPN, and thus it can conveniently be transplanted to current two-stage detectors. Since it adopts the same network architecture, the running time of DeRPN is approximately the same as RPN. In addition, DeRPN also shares convolutional layers with the second-stage detector. To be more specific, as shown in Fig. 2, the network is constituted with a $3\times3$ convolutional layer, followed by two sibling $1\times1$ convolutional layers (for classification and regression). The number of terms in geometric progression $\{a_n\}$ is denoted as $N$. Since the width and height employ independent anchor strings, the classification layer predicts $2\times2N$ scores to estimate whether an anchor string is matched or not matched with object edges. Besides, each anchor string predicts a width segment or height segment with the coordinates of $(x, w)$ or $(y, h)$, respectively. Therefore, the regression layer also predicts $2\times2N$ values.

\subsubsection{Scale-sensitive loss function}
The distribution of object scale is often imbalanced \cite{lin2014microsoft}, and there are usually more large objects than smaller ones. If we simply mix objects together to evaluate loss, the small objects will be severely overwhelmed by the larger ones. Benefiting from the distinct scales of anchor strings, we propose a novel scale-sensitive loss function to handle objects of different scales fairly. The scale-sensitive loss function is defined as
\begin{multline}
\label{eq2}
L(\{p_i\},\{t_i\})=\sum_{j=1}^{N}\sum_{i=1}^{M}\frac{1}{\left|R_j\right|}L_{cls}(p_i,p_i^*)\cdot 1\{i\in R_j\}\\
+\lambda\sum_{j=1}^{N}\sum_{i=1}^{M}\frac{1}{\left|G_j\right|}L_{reg}(t_i,t_i^*)\cdot 1\{i\in G_j\}
\end{multline}
in which
\begin{eqnarray}
\label{eq3-4}
\!\!\!\!\!\!\!\!\!\!\!\!\!\!\!\!\!\!\!\!\!\!\!&&R_j\!=\!\{k|s_k=a_j,k\!=\!1,2,...,M\}, \\
\!\!\!\!\!\!\!\!\!\!\!\!\!\!\!\!\!\!\!\!\!\!\!&&G_j\!=\!\{k|s_k\in A,s_k=a_j,and~p_i^*=1,k\!=\!1,2,...,M\}.
\end{eqnarray}
Here, $N$ is the number of terms in geometric progression $\{a_n\}$ and $M$ is the batch size. $s$ denotes the anchor string. $p_i$ represents the predicted probability of the $i$-th anchor string in a mini-batch. The ground-truth label $p_i^*$ is set to 1 if the anchor string is positive. Otherwise, $p_i^*$ is 0. $t_i$ is a predicted vector representing the parameterized coordinates, and $t_i^*$ is the corresponding ground truth vector. $A$ is the set of aligned anchor strings. $R_j$ denotes an index set containing those anchor strings of the same scale, and $j$ is used to indicate the scale corresponding to term $a_j$ in $\{a_n\}$. Similarly, $G_j$ is an index set containing positive aligned anchor strings of the same scale. The classification loss $L_{cls}$ is a cross-entropy loss, and the regression loss $L_{reg}$ is designed as a smoothed L1 loss. $\lambda$ is a balancing parameter for $L_{cls}$ and $L_{reg}$, which is empirically set to 10.

After regression, we decode the predicted coordinates as follows:
\begin{align}
\label{5-8}
x&=x_a+w_a\times t_x, \\
y&=y_a+h_a\times t_y, \\
w&=w_a\times \exp(t_w), \\
h&=h_a\times \exp(t_h),
\end{align}
where $(x ,w)$ and $(x_a, w_a)$ are, respectively, the coordinates of the predicted width segment and the anchor string. Analogously, $(y, h)$ and $(y_a, h_a)$ are those relative to height.

For each scale, we randomly sample at most 30 positive and negative anchor strings to form a mini-batch. Note that the anchor strings of the same scale share the same weighted coefficient, which is calculated by their own number. This can effectively prevent small objects being overwhelmed by larger ones. As mentioned before, we set anchor strings as a geometric progression, and thus different scales of anchor strings are explicitly distinguished by the common ratio. Therefore, the scale-sensitive loss function can naturally be applied to DeRPN.

\section{Dimension Recombination}
By using flexible anchor strings, DeRPN can predict accurate segments, serving as the edges (widths or heights) of objects. However, the final expected region proposals are two-dimensional bounding boxes. We need to reasonably combine the width and height segments to recover the region proposals. The process is illustrated in Fig. 2 (c).

\subsubsection{Pixel-wise combination algorithm}
In this paper, we propose a pixel-wise combination algorithm. First, the predicted width and height segments are decoded according to Eqs. (13)-(16). We then consider the full set of width segments (denoted as $W$ ). We filter the width segments based on probability to pick out the top-N ones ($W_N$). For each width segment $(x, w)$ in $W_N$, we choose the top-k height segments $(y^{(k)}, h^{(k)})$ at the corresponding pixels. Then, these pairs of width and height segments determine a series of specific bounding boxes $\{(x,y^{(k)},w,h^{(k)})\}$, denoted as $B_w$. The probability of a composed bounding box is given as Eq. (17).  Similarly, we can acquire $B_h=\{(x^{(k)},y,w^{(k)},h)\}$ by repeating the above steps for the height segments. We then employ non-maximum suppression (NMS) on the union set of $B_w$ and $B_h$ with an IoU threshold of 0.7. Finally, the top-$M$ bounding boxes after NMS are regarded as region proposals. In this way, we can obtain a high recall of object instances. Although this combination method introduces some background bounding boxes, the second-stage detector can suppress them with a classifier. In addition, as shown in Eq. (17), the bounding box probability is set as the harmonic mean of width and height probability. This can dramatically pull down the box probability as long as the height or width has a rather small probability, so as to remove this bounding box. We should also note that the combination algorithms are not unique. Other algorithms can be explored to achieve a high recall rate.
\begin{equation}
p^B=\left.2\middle/\left(\frac{1}{p^W}+\frac{1}{p^H}\right)\right.
\end{equation}
\section{Experiments}
In this section, we regarded RPN as the counterpart of our experiments. To verify the adaptivity, we maintained the same hyperparameters for DeRPN throughout all of our experiments without any modifications. In addition, we used the same training and testing settings for RPN and DeRPN to guarantee a fair comparison. For a comprehensive evaluation, we carried out experiments on two kinds of tasks: general object detection and scene text detection.

\subsection{General Object Detection}
\subsubsection{Dataset}
The experimental results are reported in three public benchmarks: PASCAL VOC 2007, PASCAL VOC 2012 \cite{everingham2010pascal}, and MS COCO \cite{lin2014microsoft}. PASCAL VOC and MS COCO contain 20 and 80 classes.

\subsubsection{Region proposals evaluation}
In this subsection, we drew a comparison for region proposals between DeRPN and RPN. We selected VGG16 \cite{simonyan2014very} as our backbone, and appended DeRPN or RPN to its ``conv5'' layer. We then trained models on the union set of VOC 2007 trainval and VOC 2012 trainval (``07+12'', 16551 images). The settings of RPN, training, and testing followed that of \cite{ren2015faster}. The number of output region proposals was fixed at 300.
\begin{figure}[t]
\includegraphics[width=3.in]{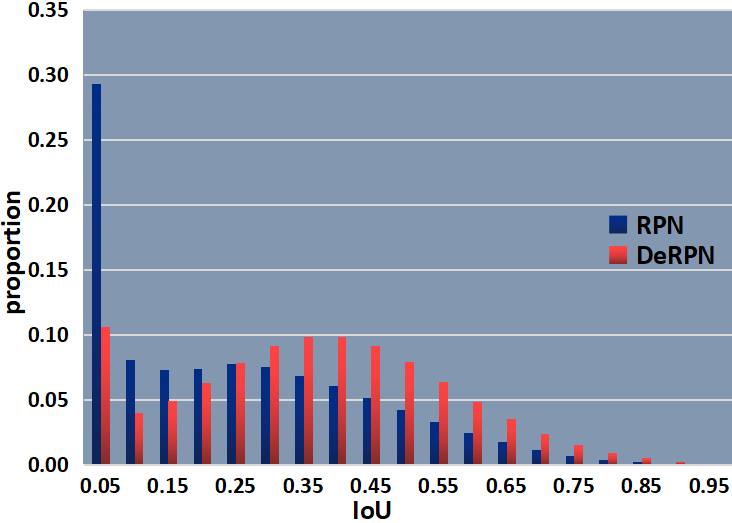}
\caption{IoU distribution of RPN and DeRPN. }
\label{predicted_vector}
\end{figure}
\begin{table}[t]
\caption{Recall rates under different IoUs, evaluated on PASCAL VOC 2007 test set.}
\label{Recall_rate}
\small
\begin{tabular}{p{0.8cm}<{\centering}p{0.6cm}<{\centering}p{0.6cm}<{\centering}p{0.6cm}<{\centering}p{0.6cm}<{\centering}p{0.6cm}<{\centering}p{0.6cm}<{\centering}p{0.6cm}<{\centering}}
\hline
{IoU} & 0.5 & 0.6 & 0.7 & 0.75 & 0.8 & 0.85 & 0.9\\
\hline
{RPN} & 96.58 & 93.19 & 81.25 &	64.83 &	38.51 &	14.83 &	3.68\\
\hline
{DeRPN} & \textbf{96.60} & \textbf{93.62} &	\textbf{85.51} &	\textbf{72.12} &	\textbf{49.01} &	\textbf{25.13} &	\textbf{9.22}\\
\hline
\end{tabular}
\end{table}
We enumerated the entire training set with trained DeRPN and RPN to collect their region proposals. Then, we made statistics of IoU distribution, which are illustrated in Fig. 3. It can be seen that the foreground (IoU$\ge$0.5) ratio of DeRPN is almost twice that of RPN, which promotes balance between the foreground and background samples since the former are usually far fewer than the latter. As demonstrated in previous studies in region scaling \cite{zhang2017feature}, hard example mining \cite{shrivastava2016training}, and sampling strategy \cite{cai2016unified}, the sample balance has a deep impact on the recall rate, which can significantly influence the training effect. Without bells and whistles, DeRPN inherently has the good property of a higher foreground ratio.
\begin{table*}[t]
\caption{Detection results on MS COCO (validation set and test set). Detector is Faster R-CNN (VGG16).}
\label{MSCOCO}
\centering
\small
\begin{tabular}{p{2cm}<{\centering}|p{2cm}<{\centering}|p{0.6cm}
<{\centering}p{0.6cm}<{\centering}p{0.6cm}<{\centering}p{0.6cm}<{\centering}p{0.6cm}<{\centering}
p{0.6cm}<{\centering}|p{0.6cm}<{\centering}p{0.6cm}<{\centering}p{0.6cm}<{\centering}
p{0.6cm}<{\centering}p{0.6cm}<{\centering}p{0.6cm}<{\centering}}
\hline
\multirow{2}{*}{Method} &\multirow{2}{*}{Anchor Type} & \multicolumn{6}{c|}{val}  & \multicolumn{6}{c}{test} \\
\cline{3-14}
&&$\rm AP $ & $\rm {AP}_{50}$ & ${\rm AP}_{75}$& ${\rm AP}_{S}$& ${\rm AP}_{M}$& ${\rm AP}_{L}$ & $\rm AP$& ${\rm AP}_{50}$& ${\rm AP}_{75}$& ${\rm AP}_{S}$& ${\rm AP}_{M}$& ${\rm AP}_{L}$\\
\hline
{RPN} & coco-type &  24.3 & 45.1 & 23.8 & 7.7 & 27.5 & 38.8& 24.4 & 45.5& 23.8 & 7.7 & 26.3 & 37.4 \\
\hline
{RPN} & voc-type & 22.9 & 42.6 & 22.2 & 5.6 & 26.2 & 37.8& 22.8 & 42.7& 22.3 & 5.6 & 24.9 & 36.3 \\
\hline
{DeRPN} & fixed  & \textbf{25.7} &\textbf{46.9} &\textbf{25.5} &\textbf{9.4} &\textbf{28.2} &\textbf{39.1} & \textbf{25.5} & \textbf{47.3}& \textbf{25.4} & \textbf{9.2} & \textbf{26.9} & \textbf{38.3} \\
\hline
\end{tabular}
\end{table*}

According to Fig. 3, we can also evaluate the mean IoU for all region proposals. The calculated mean IoUs of DeRPN and RPN are 0.34 and 0.22, respectively. This reveals that region proposals of DeRPN are more accurate and enclose objects more tightly than RPN. Furthermore, we evaluated the recall rate for DeRPN and RPN on the VOC 2007 test set ($\sim$5k images). Different IoU thresholds within [0.5, 0.9] were adopted to verify the efficacy of our method. From Table \ref{Recall_rate}, we can see that the recall rate of DeRPN surpassed that of RPN by a large margin, especially for higher IoUs within [0.7, 0.9].
\begin{table}[h]
\caption{Detection results on PASCAL VOC 2007 and 2012 test sets. Detector is Faster R-CNN. }
\label{VOC2007-2012}
\centering
\small
\begin{tabular}{p{1cm}<{\centering}|p{1.3cm}<{\centering}|p{0.95cm}<{\centering}p{0.7cm}<{\centering}|p{0.95cm}<{\centering}p{0.7cm}<{\centering}}
\hline
\multirow{2}{*}{Method}& \multirow{2}{*}{Backbone} & \multicolumn{2}{c|}{VOC 2007}  & \multicolumn{2}{c}{VOC 2012} \\
\cline{3-6}
 & & Data & mAP & Data & mAP\\
\hline
 RPN & VGG16 & 07+12 & 73.2& 07++12 & 70.4\\
DeRPN & VGG16 & 07+12 & \textbf{76.5}& 07++12 & \textbf{71.9}\\
\hline
 RPN & Res101 & 07+12 & 76.4& 07++12 & 73.8\\
DeRPN & Res101 & 07+12 & \textbf{78.8}& 07++12 & \textbf{76.5}\\
\hline
\end{tabular}
\end{table}
To sum up, our proposed DeRPN has better properties than RPN, including a higher foreground ratio, more accurate region proposals, and improved recall rate. The superiority of DeRPN benefits from its advanced dimension-decomposition mechanism, which considerably reduced the heavy burden from variant object shapes. Through flexible anchor strings, the object width and height independently seek the most appropriate regression references. After combining accurate predicted segments, DeRPN is able to produce high-quality region proposals for diverse objects.

\subsubsection{Experiments on PASCAL VOC}
In order to verify the overall improvement, DeRPN was integrated into a classical framework, Faster R-CNN (FRCN) \cite{ren2015faster}. Note that the original FRCN adopts RPN to generate region proposals. We replaced RPN with DeRPN to constitute a new detector, and then conducted comparisons. We still fixed the hyperparameters of DeRPN without any modifications. All other settings including RPN, training, and testing were maintained to be the same as FRCN. The models were implemented based on the famous networks VGG16 and ResNet-101 \cite{he2016deep}.

First, we trained our models on the set of ``07+12'' and tested them on the VOC 2007 test set. The experimental results are presented in Table \ref{VOC2007-2012}. The VGG16 result of DeRPN is 76.5\%, which is higher than the 73.2\% of RPN. Likewise, the ResNet-101 result of DeRPN surpassed that of RPN by 2.4\%. In addition, we used the union set of VOC 2007 trainval+test and V0C 2012 trainval (``07++12'', 21503 images) to train our models. The test set is VOC 2012 test (10991 images). From Table \ref{VOC2007-2012}, we can see DeRPN still retains its superiority, i.e., \textbf{71.9}\% against 70.4\% and \textbf{76.5}\% against 73.8\%.
\subsubsection{Experiments on MS COCO}
In addition, we verified our method on MS COCO 2017, which consists of a training set ($\sim$118k images), test set ($\sim$20k images) and validation set (5k images). We utilized the Faster R-CNN (VGG16) to evaluate the performance of DeRPN and RPN. It is worth noticing that, to handle small objects in MS COCO, \citeauthor{ren2015faster} redesigned the anchor boxes of RPN, changing the scales from [8,16,32] to [4,8,16,32]. We call the anchor boxes used in MS COCO the coco-type, with voc-type for PASCAL VOC. To investigate the gap between these two types of anchor boxes, we implemented RPN with coco-type and voc-type anchor boxes on MS COCO. In addition, the hyperparameters of DeRPN including the anchor strings were still unchanged, which are denoted as ``fixed''. The results are listed in Table \ref{MSCOCO}, from which we see that DeRPN still outperforms RPN on both the test and validation sets. The performance of the RPN-based detector was significantly curtailed when we used voc-type anchor boxes on MS COCO. This phenomenon demonstrates that the anchor boxes in RPN are very sensitive. Improper settings of anchor boxes will dramatically decrease the detection accuracy. Nevertheless, there is no such problem for DeRPN.
\begin{table*}[t]
\caption{Detection results on ICDAR 2013 test set. Values are expressed in recall/precision/F-measure format.}
\label{ICDAR2013}
\centering
\small
\begin{tabular}{p{1.2cm}<{\centering}|p{1.5cm}<{\centering}|p{1.8cm}<{\centering}|p{1.cm}<{\centering}|p{3cm}<{\centering}
|p{3cm}<{\centering}|p{3cm}<{\centering}}
\hline
{Method} & Proposal & Backbone & FPS & ICDAR2013 & DetEval & IoU \\
\hline
\multirow{2}{*}{FRCN} & RPN & VGG16 & 16.3 & 70.25~/~84.71~/~76.80 &	70.90~/~85.16~/~77.38 & 72.60~/~86.41~/~78.91  \\
 & DeRPN & VGG16 & 15.4 & \textbf{77.46}~/~\textbf{86.79}~/~\textbf{81.86} & \textbf{78.06}~/~\textbf{87.28}~/~\textbf{82.42} &	\textbf{79.54}~/~\textbf{89.15}~/~\textbf{84.07}  \\
\hline
\multirow{2}{*}{R-FCN} & RPN & ResNet-101 & 13.3 & 81.77~/~\textbf{92.67}~/~86.88 & 82.26~/~\textbf{93.08}~/~87.34	& 80.64~/~91.79~/~85.85  \\
 & DeRPN & ResNet-101 & 12.8 & \textbf{86.52}~/~{92.21}~/~\textbf{89.28} & \textbf{86.68}~/~{92.62}~/~\textbf{89.55} & \textbf{86.85}~/~\textbf{92.24}~/~\textbf{89.46}  \\
\hline
\multirow{2}{*}{MRCN} & RPN & ResNet-101 & 4.04 & 84.55~/~\textbf{94.67}~/~89.32 & 85.02~/~\textbf{94.69}~/~89.60 & 84.02~/~\textbf{93.69}~/~88.59 \\
 & DeRPN & ResNet-101 & 5.02 & \textbf{87.80}~/~93.71~/~\textbf{90.66}   & \textbf{88.60}~/~93.94~/~\textbf{91.19}  & \textbf{86.12}~/~92.18~/~\textbf{89.05}  \\
\hline
\end{tabular}
\end{table*}
\subsection{Scene Text Detection}
This section further verifies the adaptivity of DeRPN via scene text detection. Note that scene texts are very challenging for their complex morphology. In general, practical researches revealed that we cannot obtain satisfactory results on scene text datasets by directly using general object detectors \cite{liao2017textboxes,liu2017deep}. However, DeRPN can be directly employed on any task without any modifications for hyperparameters or specialized optimization.
\subsubsection{Dataset}
We carried out experiments on two benchmarks: ICDAR 2013 \cite{Karatzas2013ICDAR} and COCO-text \cite{veit2016coco}. ICDAR 2013 contains 229 training images and 233 test images captured from real-world scene. Besides, we append another 4k images that we gathered to enrich the training set. As for the COCO-Text dataset, it is currently the largest dataset for scene text detection. This dataset is based on MS COCO and contains 43,686 training images, 10,000 validation images, and 10,000 test images.
\subsubsection{Experiments on ICDAR 2013}
We employed FRCN, R-FCN \cite{dai2016r}, and Mask R-CNN \cite{he2017mask} (MRCN) to evaluate our method. As MRCN adopted Feature Pyramid Network (FPN) \cite{lin2016feature}, we appended DeRPN on pyramid levels of $\{P_3,P_4,P_5\}$ in FPN. There are three different evaluation protocols for ICDAR 2013: DetEval, ICDAR 2013 evaluation, and IoU evaluation. Under different protocols, the final results are presented as recall, precision and F-measure. As a trade-off, the F-measure is the harmonic mean of recall and precision. From Table \ref{ICDAR2013}, it is apparent that DeRPN exhibits superior performance for the three different protocols. The F-measure improvement of DeRPN benefits from the significant increase in the recall rate. In Fig. 4, the presented detection examples also reveal that DeRPN can effectively recall small texts and accurately detect complete long texts of extreme aspect ratios. This further demonstrates that our dimension-decomposition mechanism enables DeRPN to treat variant object shapes reasonably well. We also evaluated the inference time on a single TITAN XP GPU. In Table \ref{ICDAR2013}, the FPS of DeRPN is approximately the same as that of RPN.
\begin{figure}[!t]
\centering
\includegraphics[width=3.2in]{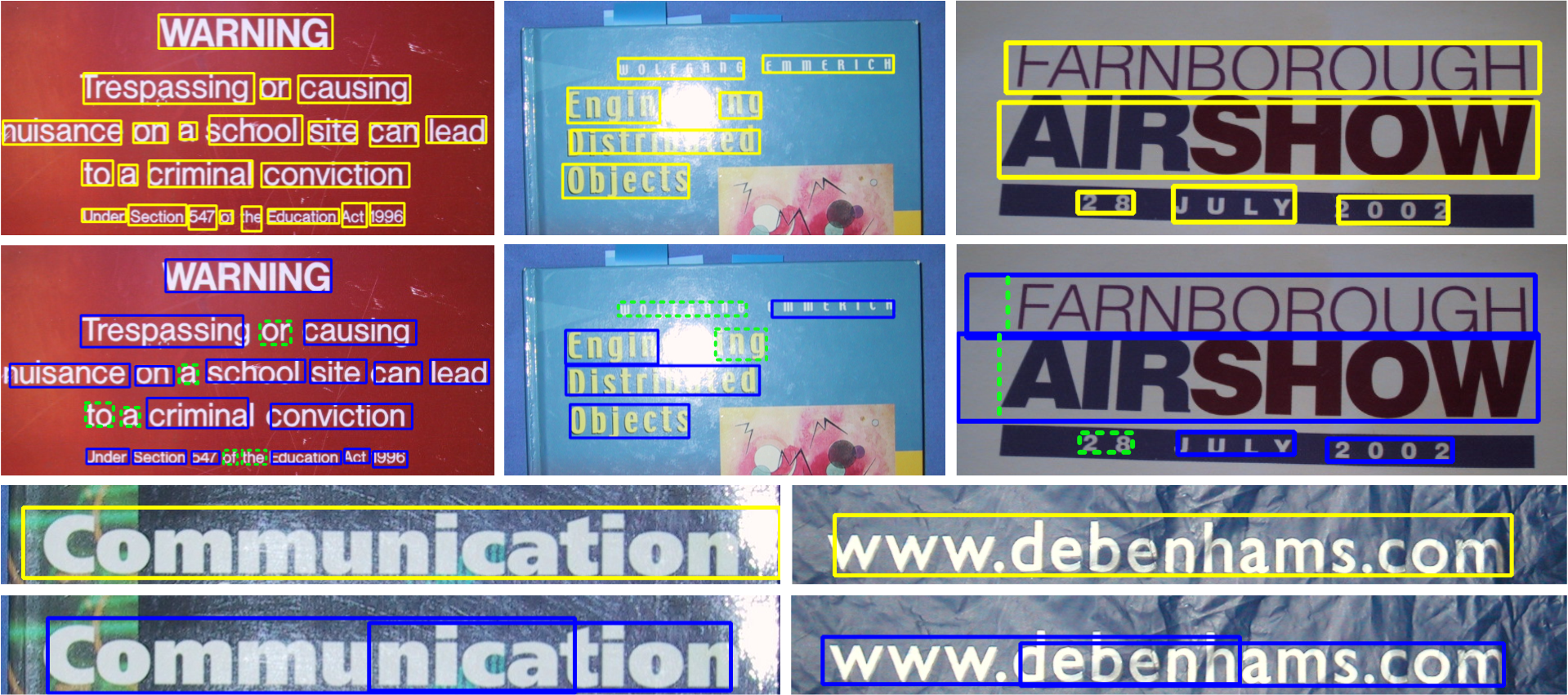}
\caption{Detection examples for ICDAR 2013 test set. Yellow boxes are detected results of DeRPN-based detector, and blue boxes are from RPN-based detector. Green dashed lines denote ground truths.}
\label{vis}
\end{figure}
\subsubsection{Experiments on COCO-Text}
We conducted experiments on COCO-Text to investigate the effects of different anchor types. We first used coco-type anchor boxes for RPN in the R-FCN baseline. In addition, we attempted to manually design anchor boxes. By imitating the methods of \cite{zhang2017feature}, we changed the aspect ratio from [0.5, 1, 2] to [0.25, 0.5, 1] because the scene texts are usually long and narrow. In addition, by following the method proposed by \cite{redmon2016yolo9000}, we enumerated the whole training set of COCO-Text to conduct k-means clustering and then produced 12 anchor boxes for RPN. The hyperparameters of DeRPN still remained unchanged.

The results in Table \ref{COCO-Text-anchor-type} were evaluated by the COCO-Text protocol. In Table \ref{COCO-Text-anchor-type}, the result is marginally worse than the baseline (coco-type) when employing manually designed anchor boxes, which reveals that it is difficult to guarantee a good result by manually setting anchor boxes for RPN. In addition, the K-means method helps increase performance to some extent. However, this improvement is still limited. When increasing the IoU to 0.75, the average precision of the K-means method is even worse than the baseline. As for DeRPN, without any modifications, it surpassed all other methods by a large margin (more than \textbf{10\%}).

Furthermore, we compared DeRPN with other specific scene text detectors including \cite{zhou2017east}, \cite{lyu2018multi}, \cite{TextBoxes2018} and so on. As these methods are evaluated by recall, precision, and F-measure, we also followed this criteria to verify our method. All listed results in Table \ref{COCO-Text STOTA} are under the constraints of single-scale testing and single model. From Table \ref{COCO-Text STOTA}, we can see that DeRPN achieved a highest F-measure of 57.11\% and established a new state-of-the-art result for COCO-Text. Unlike the other listed methods, DeRPN is designed for general object detection, which means DeRPN does not employ specialized optimization for scene text detection. However, owing to its adaptivity, DeRPN still applies to scene text detection and even outperforms specific scene text detectors.
\begin{table}[h]
\caption{Effects of anchor types, evaluated on COCO-Text test set. These methods are based on R-FCN (ResNet-101).  }
\label{COCO-Text-anchor-type}
\centering
\small
\begin{tabular}{p{1cm}<{\centering}|p{1.5cm}<{\centering}|p{0.75cm}<{\centering}
|p{0.75cm}<{\centering}|p{0.75cm}<{\centering}|p{0.75cm}<{\centering}}
\hline
{Method} & Anchor Type & AP 50 & Recall 50 & AP 75 & Recall 75 \\
\hline
{RPN} & coco-type & 40.46 &	56.76 &	13.67 &	23.43\\
\hline
{RPN} & manually & 39.76 &	56.64 &	13.49 &	22.62\\
\hline
{RPN} & K-means & 42.59 & 62.21 & 13.19 & 24.42\\
\hline
{DeRPN} & fixed & \textbf{53.43} & \textbf{73.74} & \textbf{16.99} & \textbf{30.14}\\
\hline
\end{tabular}
\end{table}
\begin{table}[h]
\caption{Comparison of specific scene text detectors. Results were evaluated on COCO-Text test set. Baseline results are given by \cite{veit2016coco}.}
\label{COCO-Text STOTA}
\small
\centering
\begin{tabular}{p{3.0cm}<{\centering}|p{0.8cm}<{\centering}|p{1.15cm}<{\centering}|p{1.5cm}<{\centering}}
\hline
{Method} & Recall & Precision & F-measure \\
\hline
{Baseline A \shortcite{veit2016coco}} & 23.30 & {83.78} & 36.48\\
\hline
{Baseline B \shortcite{veit2016coco}} & 10.70 & \textbf{89.73} & 19.14\\
\hline
{Baseline C \shortcite{veit2016coco}} & 4.70 & 18.56 & 7.47\\
\hline
{Yao et al. \shortcite{yao2016scene}} & 27.10 & 43.20  & 33.30 \\
\hline
{WordSup \shortcite{hu2017wordsup}} & 30.90 & 45.20 & 36.80 \\
\hline
\citeauthor{he2017single} (2017) & 31.00 & 46.00  & 37.00 \\
\hline
\citeauthor{lyu2018multi} (2018) & 26.20  & {69.90}  & 38.10 \\
\hline
{EAST \shortcite{zhou2017east}} & 32.40  & 50.39  & 39.45 \\
\hline
{TextBoxes++ \shortcite{TextBoxes2018}} & \textbf{56.00} & 55.82 & 55.91 \\
\hline
{R-FCN+DeRPN (ours)} &  55.71 &	58.58 &	\textbf{57.11}\\
\hline
\end{tabular}
\end{table}
\section{Conclusion}
In this paper, we have proposed a novel DeRPN to improve the adaptivity of detectors. Through an advanced dimension-decomposition mechanism, DeRPN can be employed directly on different tasks, datasets, or models without any modifications for hyperparameters or specialized optimization. Comprehensive evaluations demonstrated that the proposed DeRPN can significantly outperform the previous RPN. To our best knowledge, DeRPN is the first method that achieves outstanding performances in both general object detection and scene text detection without any tuning.

In the future, we will try to apply the dimension-decomposition mechanism to one-stage detectors and then improve their adaptivity.

\section{Acknowledgement}
This research is supported in part by GD-NSF (no.2017A030312006), the National Key Research and Development Program  of China (No. 2016YFB1001405),  NSFC (Grant No.: 61472144, 61673182, 61771199), and GDSTP (Grant No.:2017A010101027), GZSTP (no. 201607010227).

\bibliographystyle{aaai}
\bibliography{my_paper}
\end{document}